%% file: main.tex
\documentclass{article}
\usepackage{ijcai25}

\usepackage{times}
\usepackage{soul}
\usepackage{url}
\usepackage[hidelinks]{hyperref}
\usepackage[utf8]{inputenc}
\usepackage[small]{caption}
\usepackage{graphicx}
\usepackage{amsmath}
\usepackage{amsthm}
\usepackage{booktabs}
\usepackage[switch]{lineno}

\usepackage{physics}

\usepackage{amssymb}
\usepackage{xcolor}
\usepackage{multirow}

\usepackage[ruled,vlined,linesnumbered]{algorithm2e}
    
    \SetCommentSty{mycommfont}

\usepackage{subcaption}
\input{macro}

\title{Counterfactual Explanations for Continuous Action Reinforcement Learning}

\author{
Shuyang Dong$^1$
\and
Shangtong Zhang$^1$\And
Lu Feng$^1$\\
\affiliations
$^1$University of Virginia\\
\emails
\{sd3mn, shangtong, lu.feng\}@virginia.edu
}

\begin{document}

\maketitle

\begin{abstract}
    \input{0_abstract}
\end{abstract}

\section{Introduction} \label{sec:intro} 
\input{1_intro}

\section{Related Work} \label{sec:related} 
\input{2_related}

\section{Problem Formulation} \label{sec:problem} 
\input{3_problem}

\section{Approach} \label{sec:approach} 
\input{4_approach}

\section{Experiments} \label{sec:exp} 

\input{5_exp}

\section{Conclusion} \label{sec:conclu} 
\input{6_conclu}

\section*{Acknowledgements}
This work was supported in part by the U.S. National Science Foundation under Grants CCF-1942836 and CCF-2131511. The opinions, findings, conclusions, or recommendations expressed in this material are those of the author(s) and do not necessarily reflect the views of the sponsoring agencies.

\bibliographystyle{named}
\bibliography{references}

\appendix


\end{document}

%% file: macro.tex
\newcommand{\sectref}[1]{Section~\ref{#1}}

\newcommand{\figref}[1]{Figure~\ref{#1}}
\newcommand{\tabref}[1]{Table~\ref{#1}}

\newcommand{\eqnref}[1]{Equation~\ref{#1}}

\newcommand{\probref}[1]{Problem~\ref{#1}}

\newcommand{\agref}[1]{Algorithm~\ref{#1}}

\newcommand{\figfigref}[2]{Figures~\ref{#1} and \ref{#2}}

\newcommand{\tabtabref}[2]{Tables~\ref{#1} and \ref{#2}}

\makeatletter
\@addtoreset{theorem}{section}
\makeatother

\newtheorem{problem}{Problem}

\newcounter{exampcount}
\DeclareMathOperator*{\argmin}{arg\,min}

\def\cT{\mathcal{T}}
\def\cB{\mathcal{B}}
\def\exp{\mathbb{E}}
\def\adv{\mathsf{adv}}

\def\exp{{\mathbb{E}}}

%% file: 0_abstract.tex
Reinforcement Learning (RL) has shown great promise in domains like healthcare and robotics but often struggles with adoption due to its lack of interpretability. Counterfactual explanations, which address ``what if” scenarios, provide a promising avenue for understanding RL decisions but remain underexplored for continuous action spaces. We propose a novel approach for generating counterfactual explanations in continuous action RL by computing alternative action sequences that improve outcomes while minimizing deviations from the original sequence. Our approach leverages a distance metric for continuous actions and accounts for constraints such as adhering to predefined policies in specific states. Evaluations in two RL domains, Diabetes Control and Lunar Lander, demonstrate the effectiveness, efficiency, and generalization of our approach, enabling more interpretable and trustworthy RL applications.

%% file: 1_intro.tex
Reinforcement Learning (RL) has shown significant potential in tackling complex decision-making tasks across diverse fields, including healthcare~\cite{yu2021reinforcement} and robotics~\cite{tang2024deep}. However, its adoption in high-stakes applications is often hindered by a critical barrier: the lack of interpretability. Understanding an RL agent’s decision-making process is essential for fostering trust and enhancing performance. While there is a growing body of work on explainable RL~\cite{milani2023explainable}, most methods focus on summarizing policies or explaining individual actions using natural language or saliency maps. Counterfactual explanations, which address ``what if” questions to improve interpretability, remain underexplored in RL.

In contrast, most research on counterfactual explanations has focused on supervised learning, aiming to identify minimal changes to input features that yield a desired classifier output~\cite{verma2024counterfactual}. These approaches often incorporate constraints such as \emph{validity} (ensuring counterfactuals belong to the target class), \emph{proximity} (minimizing deviations from the original instance), \emph{actionability} (restricting changes to modifiable features), \emph{sparsity} (minimizing the number of changes), \emph{data manifold closeness} (ensuring realism by adhering to the training data distribution), and \emph{causality} (preserving known causal relationships). However, as noted in~\cite{gajcin2024redefining}, these methods cannot be directly applied to RL due to unique challenges, such as the temporal dependencies in decision-making and outcomes.

Existing research on counterfactual explanations for RL remains limited and often targets discrete action spaces. These approaches typically rely on heuristics~\cite{amitai2024explaining} or structural causal models~\cite{tsirtsis2021counterfactual} to generate counterfactual actions, or they focus narrowly on generating alternative state features~\cite{olson2021counterfactual}. A detailed discussion of these prior works is provided in \sectref{sec:related}.

To address these gaps, we propose a novel approach for generating counterfactual explanations in RL tailored for continuous action spaces. Our method computes an alternative sequence of actions for a given observed trajectory, aiming to achieve better outcomes (i.e., higher cumulative rewards) while minimizing deviations from the original actions. This is accomplished using a distance metric designed for continuous action sequences. Additionally, we extend the problem formulation to account for scenarios where actions in specific constrained states must adhere to a predefined policy.

A compelling example is the application of RL for managing blood glucose levels in diabetes patients~\cite{tejedor2020reinforcement}. Counterfactual explanations in this scenario can answer questions like: “What alternative insulin dosages could have resulted in better glycemic control?” These explanations must ensure minimal deviation from the original treatment plan while adhering to constraints, such as following a doctor’s prescription (e.g., administering specified insulin doses when glucose levels are within defined ranges). This example highlights the practical relevance of our approach in high-stakes decision-making tasks.

Instead of generating counterfactuals for individual observed trajectories one at a time, our approach computes a counterfactual policy that simultaneously generates desired counterfactuals for a set of observed trajectories (e.g., a patient’s historical glycemic control trajectories over a specific period). To enhance interpretability, the counterfactual policy is deterministic (e.g., ensuring that a specific insulin dosage, rather than a probabilistic range, is recommended at a time).

Our approach extends the Twin Delayed Deep Deterministic Policy Gradient (TD3) algorithm~\cite{fujimoto2018addressing}, introducing novel mechanisms for generating counterfactual trajectories in continuous action spaces and a sparse reward-shaping framework to balance reward gains with minimal action deviations.

Finally, we evaluated our approach in two RL domains: diabetes control using the FDA-approved UVA/PADOVA simulator~\cite{man2014uva} and Lunar Lander from OpenAI Gym~\cite{brockman2016openai}. Experimental results demonstrate the effectiveness, efficiency, and generalization of our approach, paving the way for more interpretable and trustworthy RL applications in high-stakes settings.

%% file: 2_related.tex
Existing work on counterfactual explanations has predominantly focused on supervised learning, with early efforts like \cite{wachter2018counterfactual} framing the problem as an optimization task to identify minimal changes to input features that result in a desired classifier output. 
Following this foundational work, extensive research has explored incorporating constraints to ensure counterfactuals are actionable and realistic, as surveyed in~\cite{verma2024counterfactual,guidotti2022counterfactual}. 
\cite{chen2024explain} employed RL-based methods to generate optimal counterfactual explanations for classifiers and subsequently distilled decision trees to explain the counterfactual generation process. 

Research on counterfactual explanations for RL is relatively limited compared to supervised learning. 
\cite{olson2021counterfactual} investigated counterfactual state explanations for deep RL agents in visual environments like Atari, examining how minimal changes to game images could alter an agent’s action. 
\cite{gajcin2024redefining} extended counterfactual explanation concepts from classifiers to RL, defining various types to address questions such as ``Had state features taken different values, action $a'$ (or policy $\pi'$) would be chosen instead of action $a$ (or policy $\pi$)" and ``Had the agent followed a different goal (or preferred a different objective) in state $s$, action $a’$ (or policy $\pi’$) would be chosen instead of action $a$ (or policy $\pi$)." 
Conversely, our work focuses on computing counterfactual explanations to explore what alternative actions or policies the agent could take to achieve better outcomes.

Our definition of counterfactual explanations is inspired by~\cite{tsirtsis2021counterfactual}, which identifies alternative action sequences differing in at most $k$ actions to achieve better outcomes in finite-horizon MDPs. However, their approach is limited to discrete action spaces and relies on a Gumbel-Max structural causal model. In contrast, our work targets RL with continuous actions, introduces a novel distance measure for action sequences, and avoids dependence on causal models.

More recently, \cite{amitai2024explaining} proposed a method for visually comparing an agent’s chosen action to counterfactual alternatives, conducting user studies in the highway environment with videos illustrating counterfactual outcomes. However, their approach is limited to discrete action spaces, with counterfactual actions selected heuristically. 
Instead, our work introduces an optimization-based method to compute counterfactual actions in continuous spaces, ensuring minimal distance from observed actions.

Additionally, other studies have explored counterfactual reasoning to enhance RL techniques. \cite{bica2021learning} modeled expert decisions by defining reward functions based on preferences for ``what if’’ outcomes and incorporating counterfactual reasoning into batch inverse RL. \cite{frost2021explaining} generated counterfactual trajectories by guiding agents to diverse, unseen states, enabling users to view rollouts of agent behavior and gain insights into its actions under test-time conditions. 
Our work has a different goal and formalizes counterfactual reasoning in RL through a novel problem formulation tailored for continuous action spaces.

%% file: 3_problem.tex
We consider a problem setup of an RL agent interacting with an environment modeled as a Markov decision process (MDP), denoted as $\mathcal{M} = (S, A, P, R)$, where $S$ represents the state space, $A$ denotes the (continuous) action space, $P$ is the (unknown) probabilistic transition function, and $R$ is the reward function.
At each time step $t$, the agent selects an action $a_t$ based on the current state $s_t$, receives a reward $R(s_t, a_t)$, and transitions to the next state $s_{t+1} \sim P(s_t, a_t)$ as governed by the environment dynamics. 
The agent's execution over a finite number of $n$ steps yields a trajectory
$\tau = \left\{(s_{t+i}, a_{t+i})\right\}_{i=0}^n$ 
with cumulative reward $G(\tau) = \sum_{i=0}^n R(s_{t+i}, a_{t+i})$.

Given an observed trajectory $\tau = \{(s_{t+i}, a_{t+i})\}_{i=0}^n$, we define a \emph{(positive) counterfactual trajectory} $\tau'$ as one that originates from the same initial state $s_t$ but achieves a higher cumulative reward, i.e., $G(\tau') > G(\tau)$, 
by adopting an alternative sequence of actions $\alpha(\tau') = \{a'_{t+i}\}_{i=0}^n$.
%
We define the distance between two action sequences $\alpha(\tau)$ and $\alpha(\tau')$ as:
\begin{equation}\label{eqn:distance}
D\bigl(\alpha(\tau), \alpha(\tau')\bigr) = \sum\limits_{i=0}^{n} \frac{|a_{t+i} - a'_{t+i}|}{|a_{t+i}| + \delta}
\end{equation}
where $\delta > 0$ is a parameter to ensure numerical stability. For multi-dimensional action spaces, this equation is generalized using $\ell_p$ norms.

We formulate an optimization problem to compute a counterfactual trajectory with minimal action distance.
\begin{problem} \label{p1}
Given a trajectory $\tau = \{(s_{t+i}, a_{t+i})\}_{i=0}^n$, find a counterfactual trajectory $\tau'$ starting from $s_t$ that satisfies:
\[
\argmin_{\tau'} D\bigl(\alpha(\tau), \alpha(\tau')\bigr), \quad \text{s.t.} \ G(\tau') > G(\tau).
\]
\end{problem}

Next, we consider a variant where actions in certain constrained states (denoted by $S^c \subseteq S$) must follow a predefined policy $\pi^c$, while actions in unconstrained states can be freely optimized to minimize the action distance and satisfy the reward constraint.
\begin{problem} \label{p2}
Given a trajectory $\tau = \{(s_{t+i}, a_{t+i})\}_{i=0}^n$, find a counterfactual trajectory $\tau'$ starting from $s_t$ that satisfies:
\begin{align*}
    & \argmin_{\tau'} D\bigl(\alpha(\tau), \alpha(\tau')\bigr), \\
    \text{s.t.} \quad & G(\tau') > G(\tau), \\
    & \forall s'_{t+i} \in S^c, \ a'_{t+i} \sim \pi^c(s'_{t+i}).
\end{align*}
\end{problem}

\begin{figure}[t] 
\centering \includegraphics[width=1.0\linewidth]{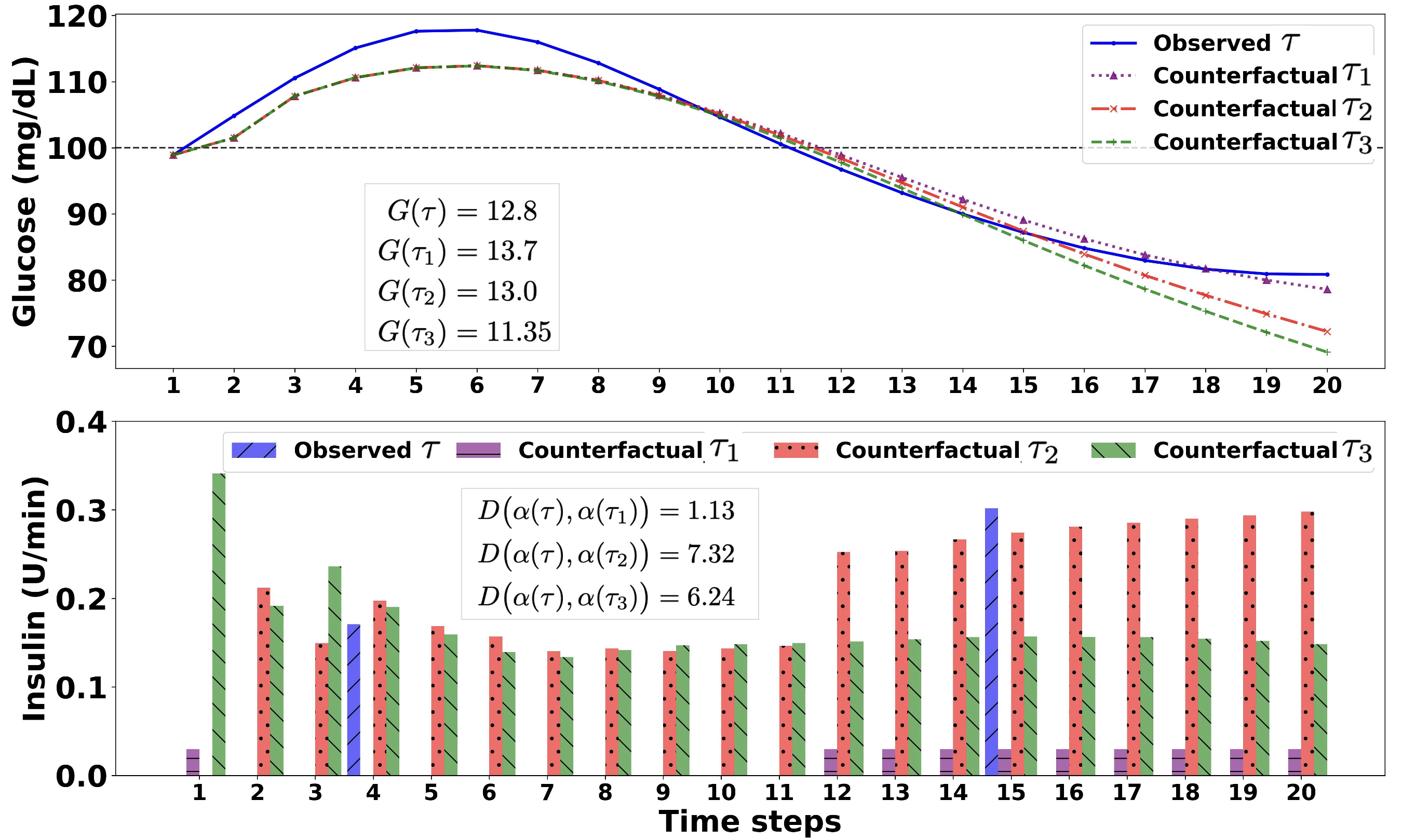} 
\caption{Observed and counterfactual trajectories of glucose levels (states) and insulin dosages (actions) over one hour.} 
\label{fig:example} 
\end{figure}

\paragraph{Motivating example.}
We illustrate these problems using the example of blood glucose (BG) control for Type 1 Diabetes patients. The state space represents BG levels, influenced by factors such as patient physiology and behaviors (e.g., eating, exercise). The action space consists of insulin doses. At each time step (e.g., every three minutes), the agent computes an insulin dosage to regulate BG levels. 
The reward function assigns positive rewards for BG levels within the target range and penalties for hypoglycemia (BG $<$ 70 mg/dL) or hyperglycemia (BG $>$ 180 mg/dL).
\figref{fig:example} shows an observed trajectory of a patient’s BG level and insulin dosage over an hour, with a cumulative reward $G(\tau) = 12.8$.

The goal is to find a counterfactual explanation addressing the question: ``What alternative treatment plan (i.e., insulin dosage sequence) could lead to better glycemic control outcomes (i.e., higher cumulative reward)?'' The counterfactual explanation should involve minimal changes from the original treatment plan (i.e., minimal distance of actions) and may include constraints (e.g., following clinical prescriptions). 

\figref{fig:example} shows three counterfactual trajectories $\tau_1$, $\tau_2$ and $\tau_3$. Among them, trajectory $\tau_1$ achieves the highest cumulative reward $G(\tau_1) = 13.7$ with the smallest action distance $D\bigl(\alpha(\tau), \alpha(\tau_1)\bigr) = 1.13$. We note that trajectory $\tau_1$ also adheres to the constraint of ``injecting 0.03 units of insulin when BG levels fall below 100 mg/dL".

%% file: 4_approach.tex

We propose a novel approach to solve the problems described in \sectref{sec:problem}. Our approach extends the Twin Delayed Deep Deterministic Policy Gradient (TD3) algorithm~\cite{fujimoto2018addressing}, incorporating mechanisms for counterfactual generation and sparse reward shaping to account for trajectory action distance.

\subsection{Solving \probref{p1}} \label{sec:sp1}

We first relax the hard constraint $G(\tau') > G(\tau)$ in \probref{p1} into a soft penalty term, yielding the following optimization problem:
\begin{align}
\min_{\tau'} D(\alpha(\tau), \alpha(\tau')) + \lambda^{-1} (G(\tau) - G(\tau'))
\end{align}
where $\lambda$ is a hyperparameter reflecting the user's preferred weight for the penalty term.
This is equivalent to
\begin{align} \label{eqn:sample-path-opt}
    \max_{\tau'} G(\tau') - \lambda D(\alpha(\tau), \alpha(\tau'))
\end{align}

Directly searching over all possible trajectories $\tau’$ is challenging without knowledge of the transition function. 
Observing that trajectories are generated by policies, 
we approximate the problem by:
\begin{align} \label{eqn:rl-opt-esssup}
    \max_{\mu} \quad  \text{ess sup}\left[G(\tau'(\mu)) - \lambda D(\alpha(\tau), \alpha(\tau'(\mu)))\right]
\end{align}
Here $\mu: S \to A$ denotes a deterministic policy, and $\tau'(\mu)$ represents the trajectory generated by following $\mu$ starting from $s_t$, the first state of $\tau$.
Specifically,
$\tau'(\mu) \doteq \qty{(s_{t+i}', a_{t+i}')}_{i=0}^n$,
where $s_t' \doteq s_t, a_{t+i}' \doteq \mu(s_{t+i}'), s_{t+i+1}' \sim P(s_{t+i}', a_{t+i}')$.
Notably, while $\mu$ is deterministic, $\tau'(\mu)$ remains a random variable due to the stochasticity in sampling $s_{t+i+1}'$ from the transition function.
The essential supremum (ess sup) is taken over all possible realizations of $\tau'(\mu)$. 
We use the essential supremum here because \eqnref{eqn:sample-path-opt} requires finding the single best trajectory generated by $\mu$.

Optimizing the essential supremum directly is difficult, as no existing methods are designed for this in sequential decision-making.
Instead, leveraging concentration inequalities (e.g., Hoeffding’s inequality) and large deviation theory, which indicate that deviations from the expectation are typically well-bounded, we optimize the expectation:
\begin{align} \label{eqn:rl-opt}
    \max_{\mu} \exp\left[G(\tau'(\mu)) - \lambda D(\alpha(\tau), \alpha(\tau'(\mu)))\right]
\end{align}
A policy $\mu$ that maximizes this expectation will, with high probability, achieve a large essential supremum. 
Here, $\exp\left[G(\tau'(\mu))\right]$ corresponds to the expected total rewards of the deterministic policy $\mu$.
Thus, \eqnref{eqn:rl-opt} can be framed as a standard RL problem, where the goal is to maximize both the total rewards $G(\tau'(\mu))$ and an additional sparse reward $-\lambda D(\alpha(\tau), \alpha(\tau'(\mu)))$, provided only at the end of each episode.

\input{4a_alg}

Given the deterministic nature of the policy $\mu$ and the proven success of TD3 in optimizing deterministic policies, we extend TD3 to solve this problem, as outlined in \agref{alg1}.
The algorithm starts with initializing the critic networks $Q_{\theta_1}$ and $Q_{\theta_2}$, the actor network $\pi_{\phi}$ with random parameters, and target networks as copies of the original networks to ensure training stability. A replay buffer $\mathcal{B}$ is also initialized to store counterfactual trajectories.
Given a set of observed trajectories $\cT$, the algorithm samples a trajectory $\tau = {(s_{t+i}, a_{t+i})}_{i=0}^n$ from $\cT$ and generates $N_c$ counterfactual trajectories. Each counterfactual trajectory $\tau_k$ starts from the same initial state as $\tau$ and is rolled out by selecting actions through the actor network $\pi_{\phi}$, with exploration noise added.
To incorporate the action distance objective in \probref{p1}, the distance $D\bigl(\alpha(\tau), \alpha(\tau_k)\bigr)$ is computed using \eqnref{eqn:distance}, and the reward is adjusted with a weighted function (line 12) following \eqnref{eqn:rl-opt}. Each generated counterfactual transition is stored in the replay buffer $\cB$.
The algorithm then samples a batch of transitions from $\cB$ to compute the target action $\tilde{a}$ and target value $y$. The critic networks are updated by minimizing the mean squared error between the predicted Q-values and the target value. To stabilize training, the actor network $\pi_{\phi}$ is updated less frequently, every $q$ iterations, using deterministic policy gradients. Finally, target networks are softly updated with a weighted average of the current and target parameters, further enhancing training stability.
The algorithm terminates after sampling and processing $N_o$ observed trajectories.

\subsection{Solving \probref{p2}} \label{sec:sp2}

To address \probref{p2}, we construct an augmented MDP, transforming \probref{p2} into an instance of \probref{p1}. Specifically, the augmented MDP has a state space defined as $\hat{S} \doteq S \setminus S^c$.
In this augmented MDP, given a state-action pair $(s, a)$, the transition proceeds as follows. 
The successor state is first determined using the original transition function $P$.
If the successor state does not belong to the constrained set $S^c$, it is presented to the agent immediately;
otherwise, the predefined policy $\pi^c$ is followed until the agent reaches a state outside $S^c$.

By constructing this augmented MDP, \probref{p2} in the original MDP is reduced to \probref{p1} in the augmented MDP. Consequently, we can directly apply \agref{alg1} in this augmented MDP to solve \probref{p2}.

%% file: 4a_alg.tex
\begin{algorithm}[tb]
\caption{Counterfactual Generation} \label{alg1}
\DontPrintSemicolon
Initialize critic networks $Q_{\theta_1}$, $Q_{\theta_2}$, and actor network $\pi_{\phi}$ with random parameters $\theta_1$, $\theta_2$, $\phi$\;
Initialize target networks: $\theta_1^{\prime} \gets \theta_1$, $\theta_2^{\prime} \gets \theta_2$, $\phi' \gets \phi$\;
Initialize replay buffer $\cB$\;
\For{$e = 1$ to $N_o$}{
    Sample a trajectory $\tau = \{(s_{t+i}, a_{t+i})\}_{i=0}^n$ from $\cT$ \;
    \For{$k = 1$ to $N_c$}{
        Set initial state $s'_t \gets s_t$ \;
        \For{$i = 0$ to $n$}{
            Select action with exploration noise: $a'_{t+i} \sim \pi_{\phi}(s'_{t+i}) + \epsilon$, $\epsilon \sim \mathcal{N}(0, \sigma)$ \;
            Observe reward $r'_{t+i}$ and next state $s'_{t+i+1}$ \;
            \If{$i = n$}{
                Adjust reward: $r'_{t+i} \gets r'_{t+i} - \lambda \cdot D\bigl(\alpha(\tau), \alpha(\tau_k)\bigr)$ \;
            }
            Store $(s'_{t+i}, a'_{t+i}, r'_{t+i}, s'_{t+i+1})$ in $\cB$ \;
        }
    }
    Sample mini-batch of $N$ transitions $(s, a, r, s')$ from replay buffer $\cB$ \;
    $\tilde{a} \gets \pi_{\phi'}(s') + \epsilon$, $\epsilon \sim \text{clip}(\mathcal{N}(0, \tilde{\sigma}), -c, c)$ \;
    $y \gets r + \gamma \min_{j=1,2} Q_{\theta_j^{\prime}}(s', \tilde{a})$ \;
    Update critics: $\theta_j \gets \arg\min_{\theta_j} N^{-1} \sum \big(y - Q_{\theta_j}(s, a)\big)^2$ \;

    \If{$e \bmod q = 0$}{
        Update actor $\phi$ using deterministic policy gradient:
        $\nabla_{\phi} J(\phi) = N^{-1} \sum \nabla_a Q_{\theta_1}(s, a) \big|_{a=\pi_{\phi}(s)} \nabla_{\phi} \pi_{\phi}(s)$ \;
        Update target networks:
        $\theta'_j \gets \eta \theta_j + (1 - \eta) \theta'_j, \quad \phi' \gets \eta \phi + (1 - \eta) \phi'$ \;
    } 
}
\end{algorithm}

%% file: 5_exp.tex
We implemented the proposed approach and evaluated it in two RL domains: (i) diabetes control using the FDA-approved UVA/PADOVA simulator~\cite{man2014uva}, and (ii) Lunar Lander from OpenAI Gym~\cite{brockman2016openai}. 
Our implementation\footnote{Code is available at: \url{https://github.com/safe-autonomy-lab/CounterfactualRL}}
is based on Stable-Baselines3~\cite{raffin2021stable}.

\paragraph{Data Generation.}
We trained a baseline policy using the Proximal Policy Optimization (PPO) algorithm~\cite{schulman2017proximal} to generate trajectories for training and test datasets (details specific to each domain are provided later). Our approach is RL-method agnostic and can generate counterfactuals from trajectories produced by any RL algorithm.

\paragraph{Baseline Method.}
As no existing methods generate counterfactuals for continuous action RL, we compare our approach to a naive baseline that generates counterfactuals by rolling out the trained baseline policy in evaluation mode without state constraints or additional training.

\paragraph{Metrics.}
Our evaluation focuses on the effectiveness, efficiency, and generalization of the proposed approach.
We define two metrics to assess the effectiveness. 
\begin{itemize}
    \item \emph{Positive Counterfactual Percentage} ($\rho_+$): This metric measures the percentage of test set trajectories for which at least one positive counterfactual trajectory (i.e., a trajectory with a higher cumulative reward than the observed trajectory) is identified among a fixed number of generated counterfactuals. This metric applies to both the baseline and the proposed approaches. 
    \item \emph{Advantage Counterfactual Percentage} ($\rho_{\adv}$):
    This metric quantifies the percentage of test set trajectories for which the proposed approach generates a more advantageous counterfactual than the baseline. 
    For an observed trajectory $\tau$, let $\tau^*_p$ and $\tau^*_b$ denote the best positive counterfactual trajectories with minimal action distance produced by the proposed and baseline approaches, respectively, satisfying $G(\tau^*_p) > G(\tau)$ and $G(\tau^*_b) > G(\tau)$.
    The proposed approach is considered advantageous if $\phi_G > \phi_D$, where:
    \[
    \phi_G = \frac{G(\tau^*_p) - G(\tau)}{G(\tau^*_b) - G(\tau)},
    \quad
    \phi_D = \frac{D\bigl(\alpha(\tau), \alpha(\tau^*_p)\bigr)}{D\bigl(\alpha(\tau), \alpha(\tau^*_b)\bigr)},
    \]
    indicating that reward gains outweigh additional action distance.
    $\rho_{\adv}$ is computed as the percentage of advantageous counterfactuals among all test trajectories with valid positive counterfactuals. 
\end{itemize}
Efficiency is assessed through learning curves of $\rho_+$ and $\rho_{\adv}$, illustrating performance improvements during the training of the proposed approach.
Generalization is evaluated by testing the approach on unseen trajectory datasets across diverse single- and multi-environment settings.

\subsection{Diabetes Control} \label{sec:diabetes}

\paragraph{MDP Environment.}
The FDA-approved UVA/PADOVA Simulator~\cite{man2014uva} was used to model glucose-insulin dynamics and simulate the effects of insulin delivery, carbohydrate intake, and other factors on blood glucose levels in Type 1 Diabetes patients. Virtual patient profiles were generated with varying parameters (e.g., age, weight, insulin sensitivity). The MDP state space includes glucose reading, glucose rate of change, and carbohydrate intake, while the action space consists of the insulin dosage at each time step. A reward function from~\cite{zhu2020basal} was used, providing the highest rewards for maintaining glucose levels within 90–140 mg/dL, smaller rewards for near-target ranges, and penalties for hypo- or hyperglycemic values, with increasing penalties for larger deviations.

\paragraph{Data Generation.}
The proposed approach was evaluated in two settings: single-environment (single patient) and multi-environment (population model with multiple patients).
In the single-environment setting, a baseline policy was trained on a chosen patient profile for 100,000 steps, with a learning rate of 0.0001 and a gradient step size of 50. Observed trajectories were generated using a sliding window method, with each trajectory representing a 20-step segment (corresponding to a one-hour patient execution history).
In the multi-environment setting, the baseline policy was trained on three patient profiles, with 3,000 steps per patient per round, continuing until stable performance was achieved. Other parameters were consistent with the single-environment setting. 
Both settings included 18 unique trajectories in each training and test set.

\paragraph{Experimental Setup.}
We evaluated three variants of the proposed approach: (P1) directly applying \agref{alg1} to solve \probref{p1}; (P2-base) adapting \agref{alg1} for \probref{p2} with constrained states $S^c$ (glucose levels below 100 mg/dL) and the baseline policy as $\pi^c$; and (P2-fixed) solving \probref{p2} with the same $S^c$ but using a fixed policy $\pi^c$ that injects 0.03 units of insulin per step.
For each variant, \agref{alg1} trained an RL model to generate counterfactual trajectories with a distance reward weight of $\lambda = 1$. Counterfactual trajectories were computed for each training set trajectory and stored in a buffer. After a warm-up phase, batches of 256 trajectories were sampled for model updates using a learning rate of 0.0001 and 50 gradient steps. The RL agent was tested every 400 interaction steps, generating 10 counterfactual trajectories per test trajectory via policy rollout for evaluation.
All three variants were compared to the same baseline method (rolling out the baseline policy without constrained states).

\paragraph{Results Analysis.}
We conducted seven independent trials for each method, and the results are analyzed below.

\emph{Effectiveness:}
\tabtabref{tab:pos}{tab:adv} present the $\rho_+$ and $\rho_\adv$ metrics evaluated on test performance across seven trials after training. All three methods outperform the baseline in Positive Counterfactual Percentage ($\rho_+$) across single- and multi-environment settings. In the single-environment setting, P1 and P2-fixed achieve the highest $\rho_+$, with P2-fixed slightly surpassing P1 in the multi-environment setting. For Advantage Counterfactual Percentage ($\rho_\adv$), P1 consistently demonstrates the best performance in both settings. P2-fixed also performs reliably, while P2-base moderately improves over the baseline in the multi-environment setting but fails to generate advantageous counterfactuals in the single-environment setting.

\begin{table}[t]
\centering
\begin{tabular}{l|l|r|r}
\toprule
\textbf{Domain} & \textbf{Method} & \textbf{Single-Env} & \textbf{Multi-Env} \\
\midrule
\multirow{3}{*}{Diabetes} 
 & P1 & \textbf{0.53 $\pm$ 0.01} & 0.44 $\pm$ 0.0 \\
 & P2-base & 0.47 $\pm$ 0.01 & 0.40 $\pm$ 0.02 \\
 & P2-fixed & \textbf{0.53 $\pm$ 0.01} & \textbf{0.48 $\pm$ 0.03} \\
 & Baseline & 0.44 $\pm$ 0.0 & 0.39 $\pm$ 0.0 \\
\midrule
\multirow{3}{*}{Lunar Lander} 
 & P1 & \textbf{0.81 $\pm$ 0.03} & \textbf{0.82 $\pm$ 0.02} \\
 & P2-base & 0.65 $\pm$ 0.08 & 0.54 $\pm$ 0.02 \\
 & P2-fixed & 0.31 $\pm$ 0.06 & 0.42 $\pm$ 0.03 \\
 & Baseline & 0.61 $\pm$ 0.02  & 0.50 $\pm$ 0.0 \\ 
\bottomrule
\end{tabular}
\caption{Mean and standard error of Positive Counterfactual Percentage ($\rho_+$) in single- and multi-environment settings.}
\label{tab:pos}
\end{table}

\begin{table}[t]
\centering
\begin{tabular}{l|l|r|r}
\toprule
\textbf{Domain} & \textbf{Method} & \textbf{Single-Env} & \textbf{Multi-Env} \\
\midrule
\multirow{3}{*}{Diabetes} 
 & P1 & \textbf{0.67 $\pm$ 0.0} & \textbf{0.86 $\pm$ 0.0} \\
 & P2-base & 0.0 $\pm$ 0.0 & 0.41 $\pm$ 0.02 \\
 & P2-fixed & 0.52 $\pm$ 0.03 & 0.65 $\pm$ 0.06 \\
\midrule
\multirow{3}{*}{Lunar Lander} 
 & P1 & \textbf{0.38 $\pm$ 0.03} & \textbf{0.94 $\pm$ 0.04} \\
 & P2-base & 0.36 $\pm$ 0.04 & 0.72 $\pm$ 0.05 \\
 & P2-fixed & 0.32 $\pm$ 0.09 & 0.63 $\pm$ 0.04 \\
\bottomrule
\end{tabular}
\caption{Mean and standard error of Advantage Counterfactual Percentage ($\rho_\adv$) in single- and multi-environment settings.}
\label{tab:adv}
\end{table}

\begin{figure}[t]
    \centering
    \begin{subfigure}[b]{0.38\textwidth}
        \centering
        \includegraphics[width=\textwidth]{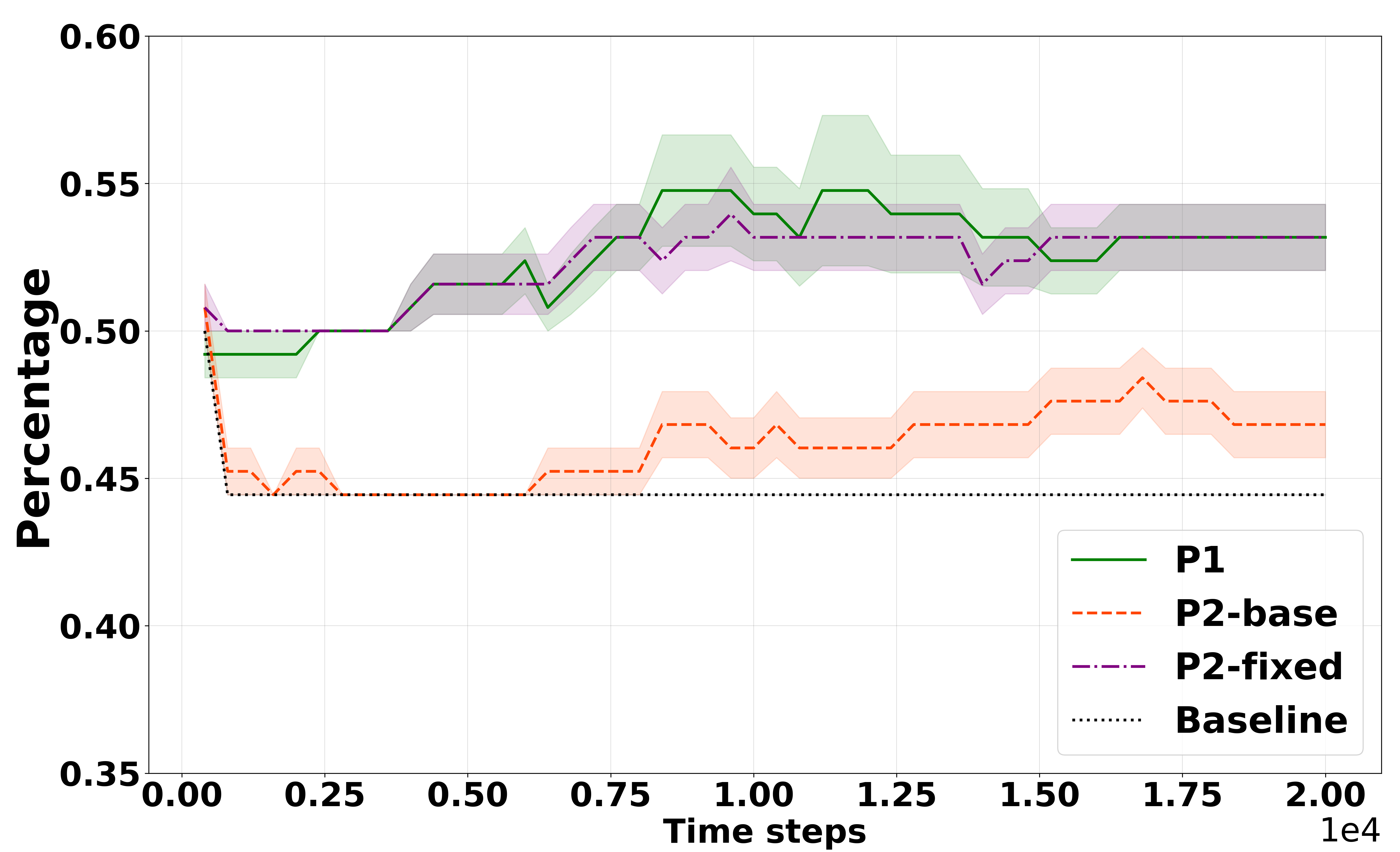} 
        \caption{Single-Environment}
        \label{fig:diabetes-pos-a}
    \end{subfigure}
    \hfill
    \begin{subfigure}[b]{0.38\textwidth}
        \centering
        \includegraphics[width=\textwidth]{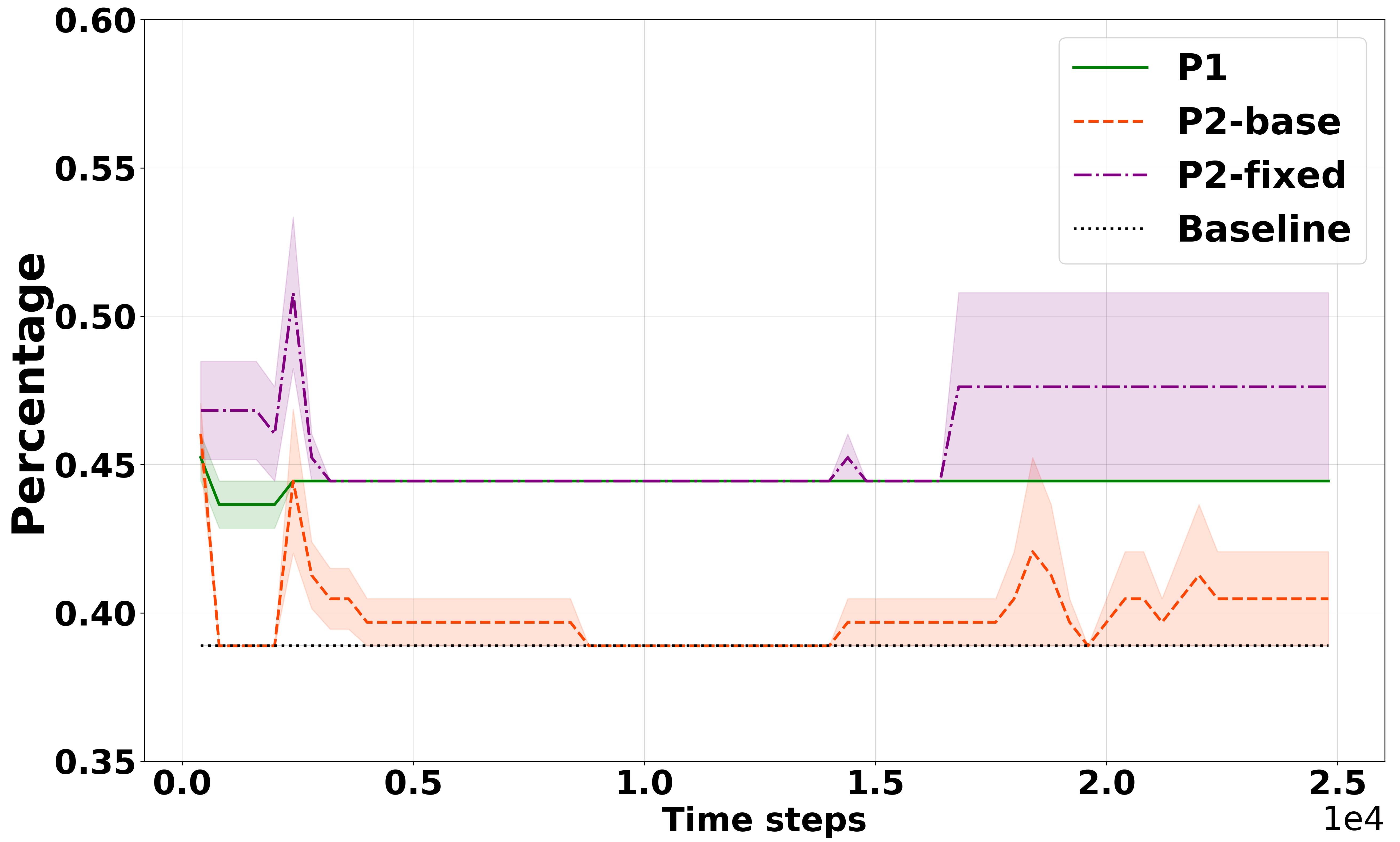} 
        \caption{Multi-Environment}
        \label{fig:diabetes-pos-b}
    \end{subfigure}
    \caption{Learning curves of Positive Counterfactual Percentage ($\rho_+$) for the Diabetes Control domain.}
    \label{fig:diabetes-pos}
\end{figure}

\emph{Efficiency:}
\figfigref{fig:diabetes-pos}{fig:diabetes-adv} show the learning curves of $\rho_+$ and $\rho_\adv$ for each method in single- and multi-environment settings, with solid lines representing the mean and shaded areas denoting the standard error across seven trials. The baseline curve in \figref{fig:diabetes-pos} represents counterfactuals generated by rolling out the baseline policy at each evaluation point, without additional training.
In the single-environment setting (\figfigref{fig:diabetes-pos-a}{fig:diabetes-adv-a}), P1 converges quickly and achieves the highest $\rho_+$ and $\rho_\adv$. P2-fixed steadily improves, matching P1 in $\rho_+$ and delivering strong $\rho_\adv$ performance. P2-base, however, converges more slowly and inconsistently, underperforming relative to P1 and P2-fixed.
In the multi-environment setting (\figfigref{fig:diabetes-pos-b}{fig:diabetes-adv-b}), P2-fixed achieves the highest $\rho_+$, while all three methods converge rapidly and maintain stable $\rho_\adv$ performance. Among them, P1 achieves the highest $\rho_\adv$, followed closely by P2-fixed, with P2-base showing slightly weaker but reliable results.

\begin{figure}[t]
    \centering
    \begin{subfigure}[b]{0.38\textwidth}
        \centering
        \includegraphics[width=\textwidth]{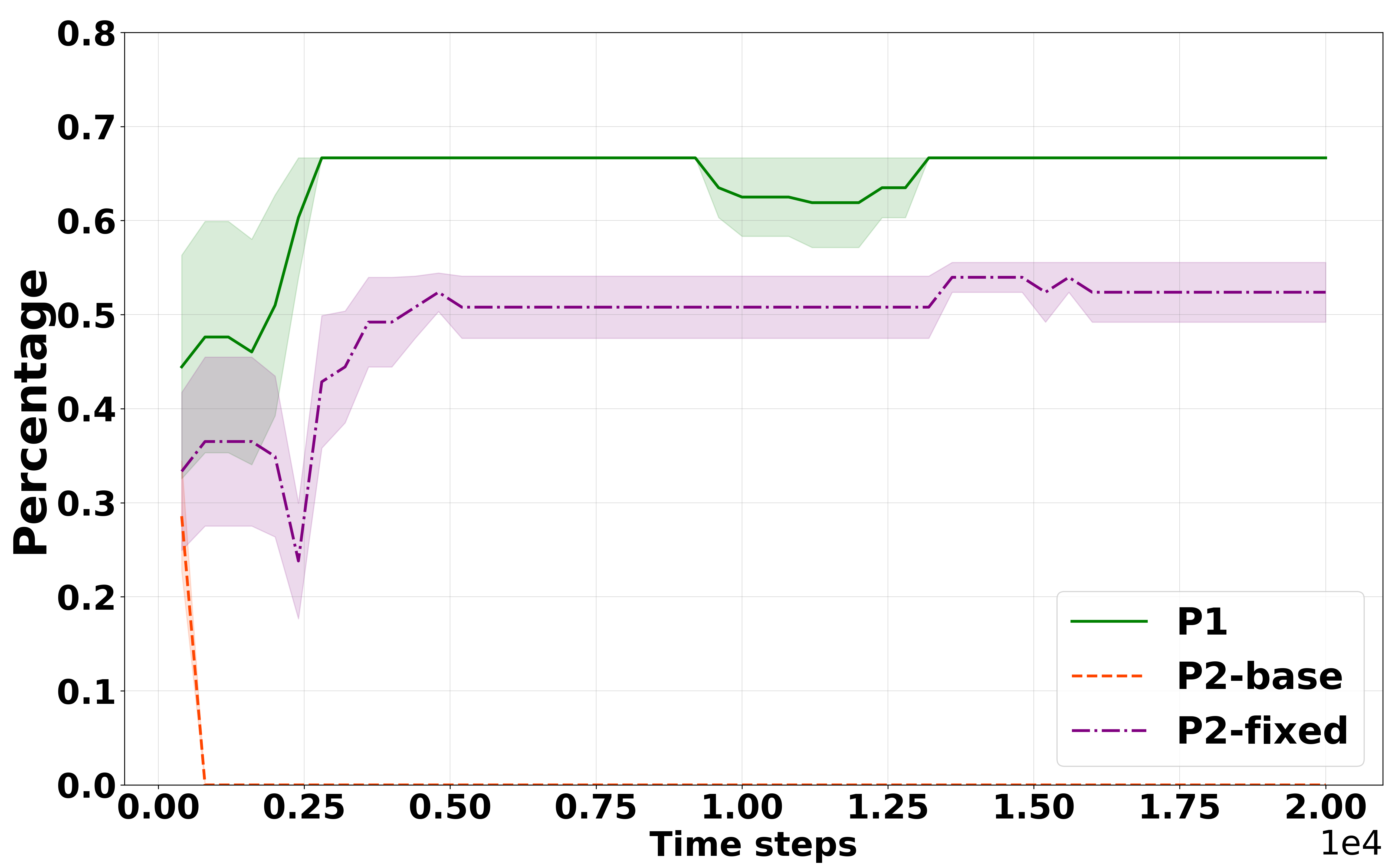} 
        \caption{Single-Environment}
        \label{fig:diabetes-adv-a}
    \end{subfigure}
    \hfill
    \begin{subfigure}[b]{0.38\textwidth}
        \centering
        \includegraphics[width=\textwidth]{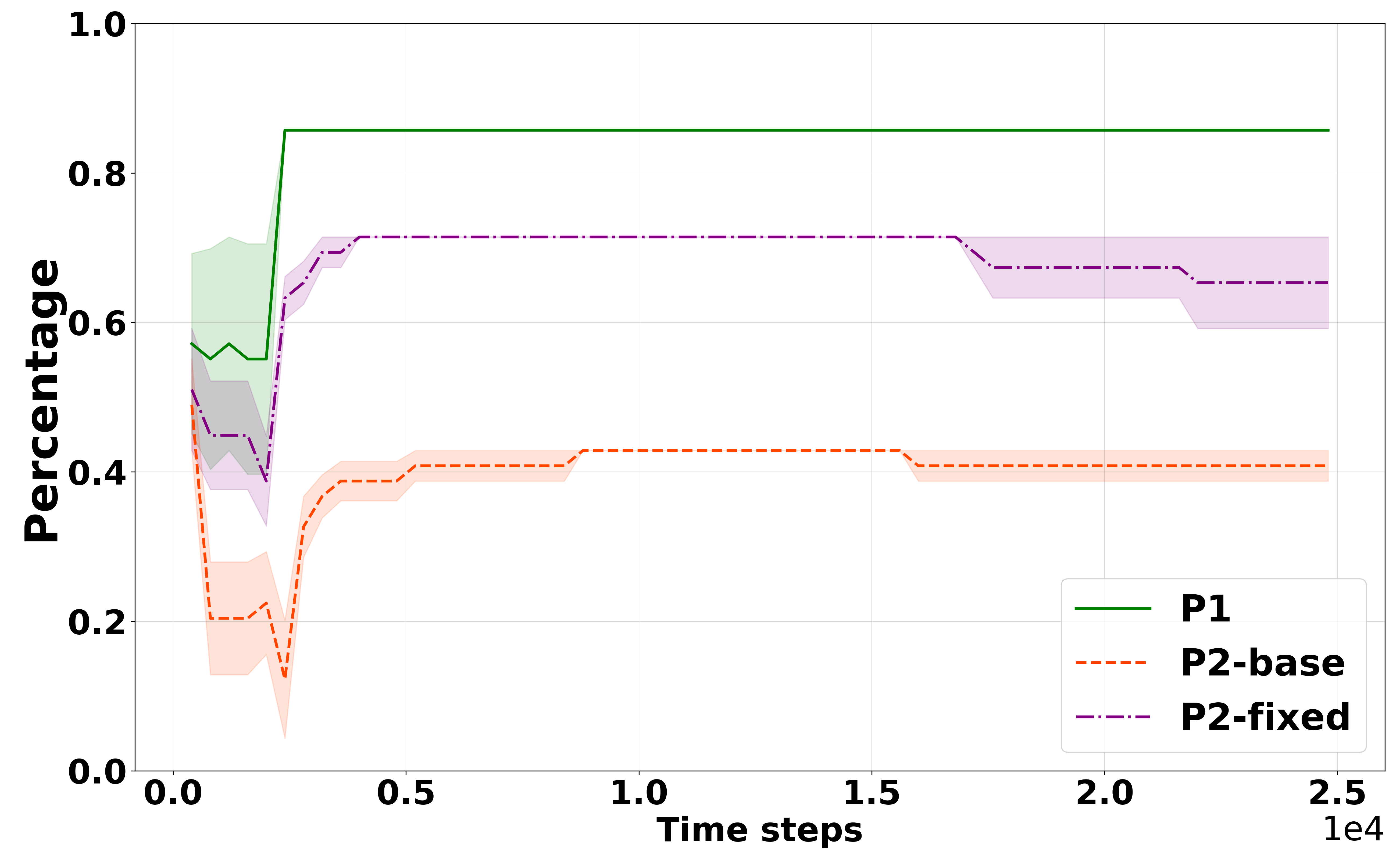} 
        \caption{Multi-Environment}
        \label{fig:diabetes-adv-b}
    \end{subfigure}
    \caption{Learning curves of Advantage Counterfactual Percentage ($\rho_\adv$) for the Diabetes Control domain.}
    \label{fig:diabetes-adv}
\end{figure}

\emph{Generalization:}  
Performance across settings highlights each method's generalization capabilities. P1 generalizes well, maintaining strong and consistent performance in both settings. P2-fixed also generalizes effectively, excelling in the multi-environment setting. P2-base shows limited generalization, struggling in the single-environment but improving slightly in the multi-environment, with notable variability.

\emph{Summary:}  
In the diabetes control experiments, the proposed methods consistently outperform the baseline in effectiveness, efficiency, and generalization. P1 is the most effective and efficient, while P2-fixed excels in generalization to multi-environment setting. P2-base, though weaker overall, demonstrates modest generalization potential.

\subsection{Lunar Lander} \label{sec:lunar}

\paragraph{MDP Environment.}  
The Lunar Lander environment from OpenAI Gym~\cite{brockman2016openai} simulates a 2D rocket attempting to land safely on a designated pad. The MDP state space includes eight continuous variables: the lander's x and y coordinates, x and y velocities, orientation angle, angular velocity, and two binary indicators for the left and right legs' contact with the ground. The action space includes two continuous variables controlling the throttle of the main engine and the lateral boosters. The reward function encourages safe landings, efficient fuel use, and minimal engine overuse. 

\paragraph{Data Generation.}
The proposed approach was evaluated under single- and multi-environment settings. In the single-environment setting, a baseline policy was trained for 3,000 steps with a learning rate of 0.0001 and a gradient step size of 20. In the multi-environment setting, the baseline policy was trained across three gravity values, with 500 steps per environment per round, until stable performance was achieved. As in the Diabetes domain, trajectories were generated using a sliding window method, with each 20-step segment forming a trajectory. Both settings included 12 randomly sampled trajectories in each training and test set.

\paragraph{Experimental Setup.} 
The proposed approach was used to generate counterfactual trajectories with a learning rate of 0.00001 and a gradient step size of 20, following a process similar to the diabetes experiments. Three variants were evaluated: P1, P2-base, and P2-fixed. In P2-base and P2-fixed, constrained states $S^c$ were defined as those with x-velocity in $[-0.18, 0.18]$. The policy $\pi^c$ was set as the baseline policy in P2-base, while P2-fixed used a fixed policy with action vectors set to $(0, 0)$.

\paragraph{Results Analysis.}
We conducted seven independent trials for each method and analyzed the results below.

\emph{Effectiveness:}
As shown in \tabtabref{tab:pos}{tab:adv}, P1 achieves the highest Positive Counterfactual Percentage ($\rho_+$) and Advantage Counterfactual Percentage ($\rho_\adv$) in both single- and multi-environment settings, surpassing all other methods and the baseline. While P2-base falls short of P1 in both metrics, it outperforms the baseline. P2-fixed shows the weakest performance, slightly below the baseline in $\rho_+$.

\begin{figure}[t]
    \centering
    \begin{subfigure}[b]{0.38\textwidth}
        \centering
        \includegraphics[width=\textwidth]{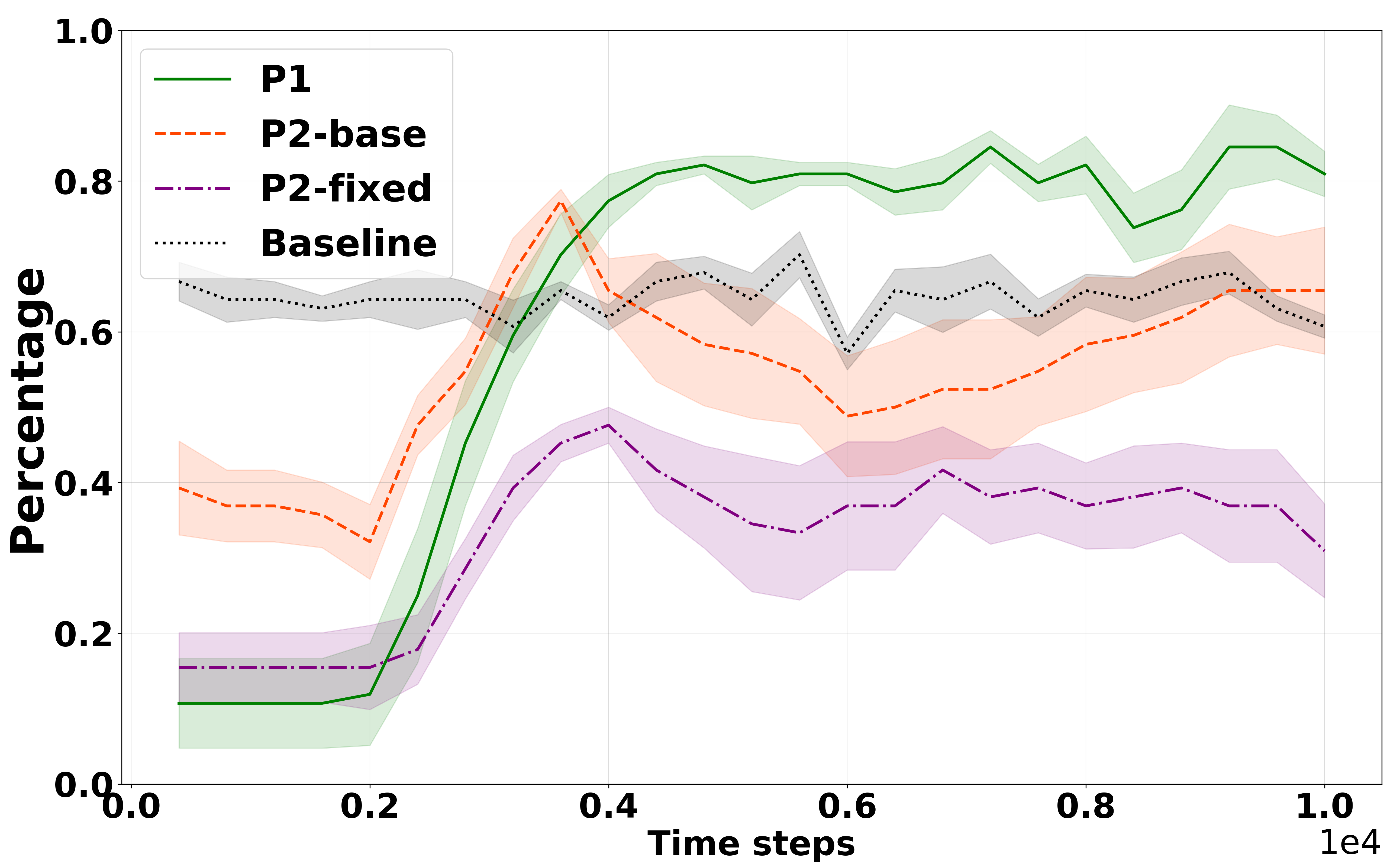} 
        \caption{Single-Environment}
        \label{fig:lunar-pos-a}
    \end{subfigure}
    \hfill
    \begin{subfigure}[b]{0.38\textwidth}
        \centering
        \includegraphics[width=\textwidth]{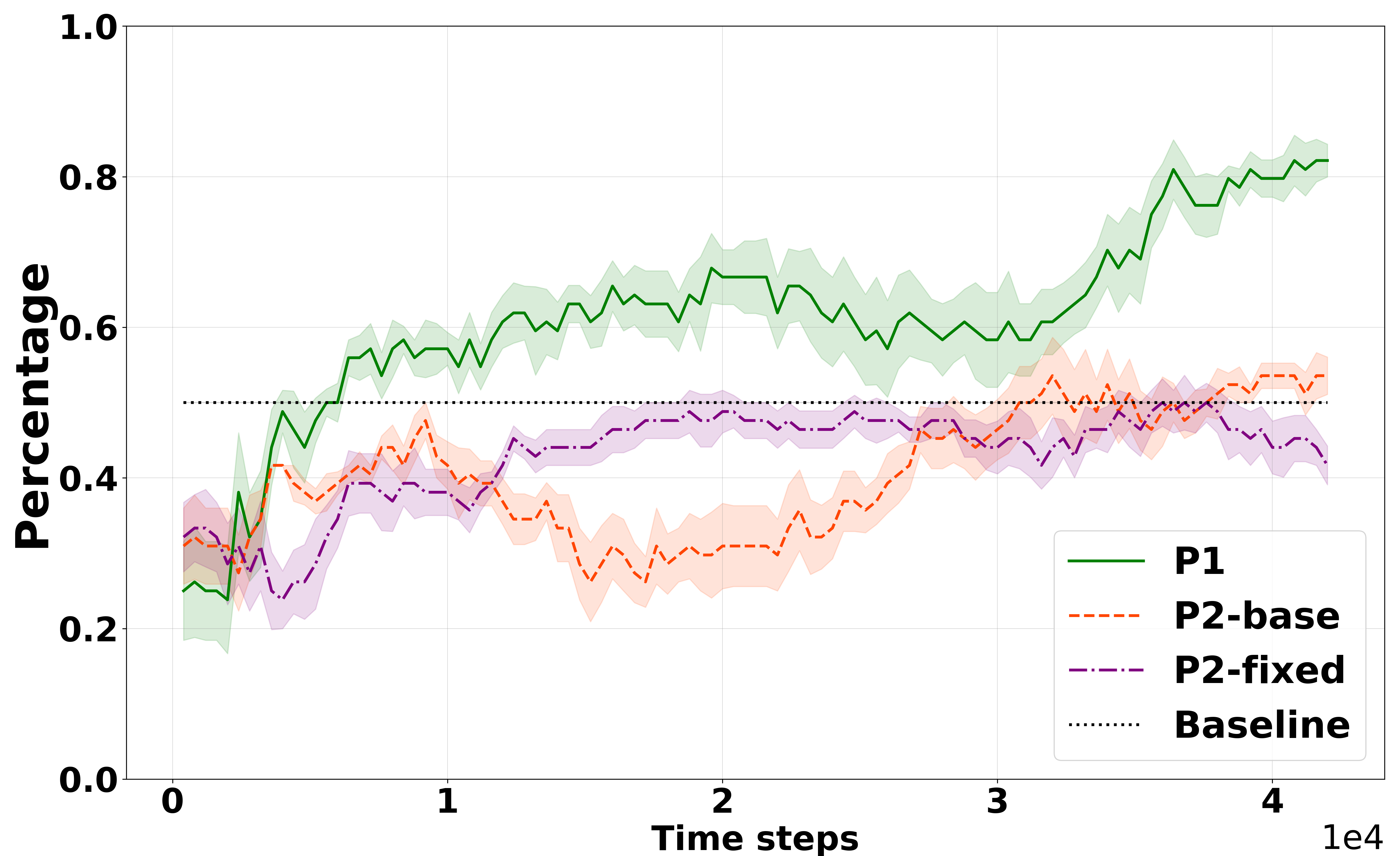} 
        \caption{Multi-Environment}
        \label{fig:lunar-pos-b}
    \end{subfigure}
    \caption{Learning curves of Positive Counterfactual Percentage ($\rho_+$) for the Lunar Lander domain.}
    \label{fig:lunar-pos}
\end{figure}

\emph{Efficiency:}
The learning curves in \figfigref{fig:lunar-pos}{fig:lunar-adv} depict the efficiency of each method in terms of Positive Counterfactual Percentage ($\rho_+$) and Advantage Counterfactual Percentage ($\rho_\adv$) for the Lunar Lander domain. In the single-environment setting (\figfigref{fig:lunar-pos-a}{fig:lunar-adv-a}), P1 converges quickly and achieves the highest $\rho_+$ and $\rho_\adv$. P2-base steadily improves, surpassing the baseline, while P2-fixed converges slowly and lags behind other methods. In the multi-environment setting (\figfigref{fig:lunar-pos-b}{fig:lunar-adv-b}), P1 consistently outperforms others in both $\rho_+$ and $\rho_\adv$, with rapid convergence and strong results. P2-base and P2-fixed demonstrate less consistent progress in $\rho_+$ but maintain competitive performance in $\rho_\adv$.

\begin{figure}[t]
    \centering
    \begin{subfigure}[b]{0.38\textwidth}
        \centering
        \includegraphics[width=\textwidth]{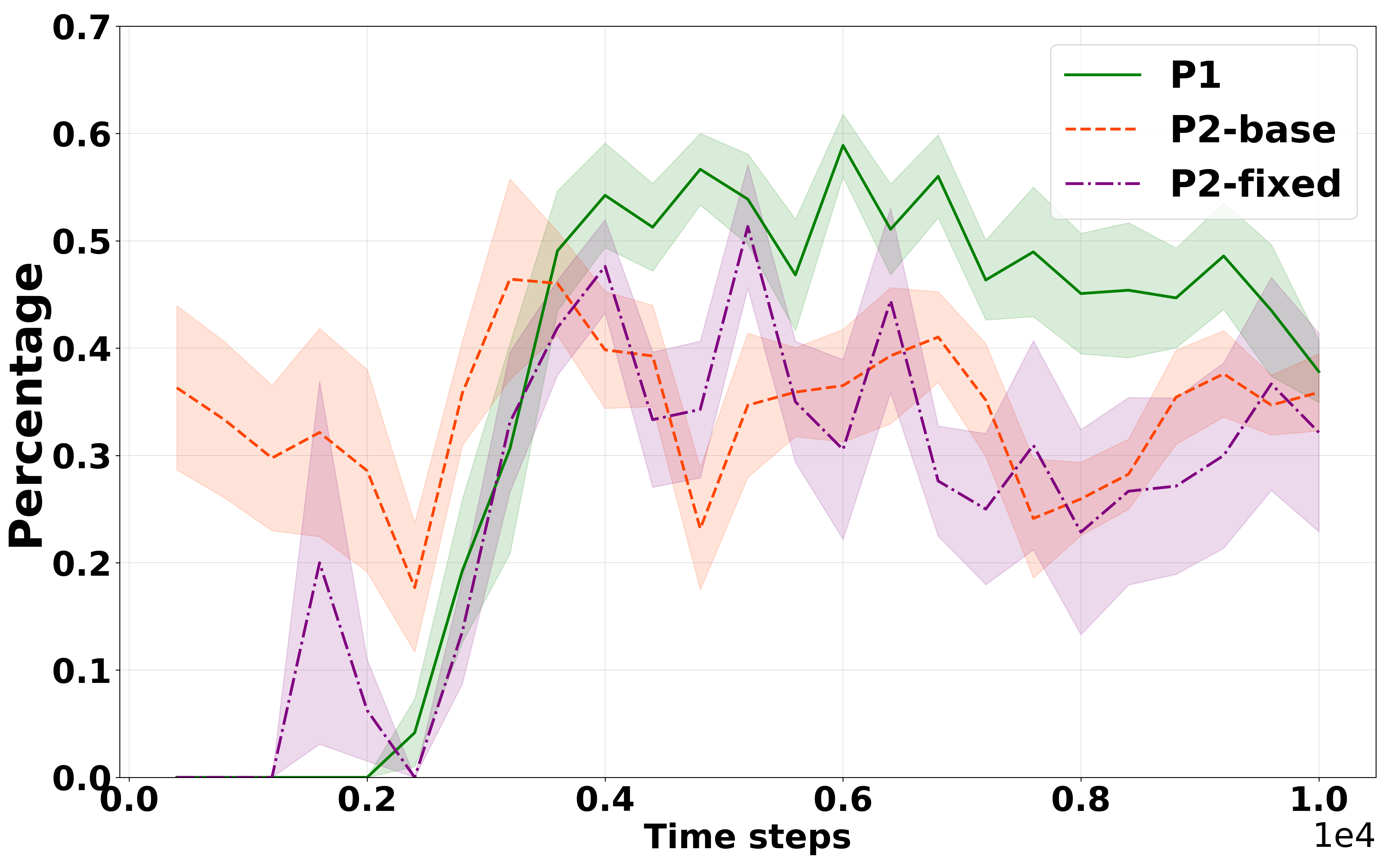} 
        \caption{Single-Environment}
        \label{fig:lunar-adv-a}
    \end{subfigure}
    \hfill
    \begin{subfigure}[b]{0.38\textwidth}
        \centering
        \includegraphics[width=\textwidth]{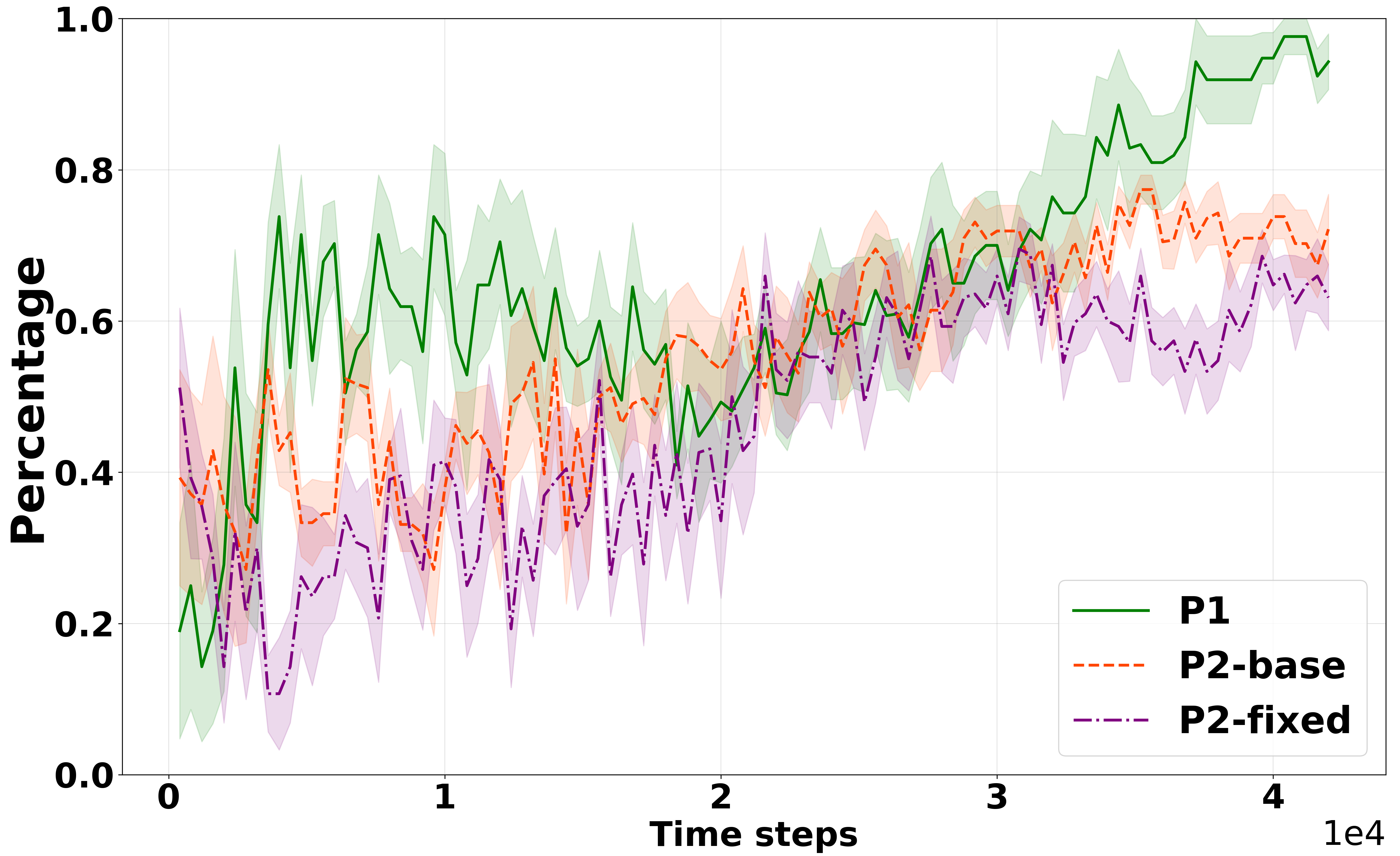} 
        \caption{Multi-Environment}
        \label{fig:lunar-adv-b}
    \end{subfigure}
    \caption{Learning curves of Advantage Counterfactual Percentage ($\rho_\adv$) for the Lunar Lander domain.}
    \label{fig:lunar-adv}
\end{figure}

\emph{Generalization:}  
All three methods generalize effectively from single- to multi-environment settings. Notably, Advantage Counterfactual Percentage ($\rho_\adv$) improves significantly in the multi-environment setting, while the learning curves remain comparable across both settings.

\emph{Summary:}  
In the lunar landing experiments, P1 consistently outperforms other methods across both settings. P2-base demonstrates moderate effectiveness and generalization, with gradual improvements and variability in performance. P2-fixed, while initially competitive, lags behind and performs worse than the baseline in some cases.

\subsection{Discussion} \label{sec:discussion}

\paragraph{Comparing Single- and Multi-Environment.}  
The $\rho_\adv$ results in multi-environment settings show marked improvement over single-environment settings across all methods and domains (\tabref{tab:adv}). This suggests that training in diverse environments enhances model performance by exposing it to a wider range of scenarios, enabling better generalization and the integration of more varied information. These findings highlight the value of tailoring counterfactual generation methods to the unique characteristics of each environment and leveraging diverse training conditions to achieve superior outcomes.

\paragraph{Comparing Results of Two Domains.}
The results from the Diabetes Control and Lunar Lander experiments highlight the influence of domain-specific characteristics on the performance of counterfactual generation methods. While P1 consistently performs best in both domains, P2-fixed outperforms P2-base in Diabetes Control, whereas P2-base surpasses P2-fixed in Lunar Lander. These differences arise from the nature of state transitions and action impacts in each environment. In Lunar Lander, actions produce immediate and short-lived effects—adjusting the engine throttle instantly changes the rocket’s velocity, with the effects quickly dissipating. Conversely, Diabetes Control involves delayed and prolonged effects, as an insulin dose takes time to affect blood glucose levels and its influence persists for an extended period. This delay likely contributes to the more stable learning curves in the Diabetes Control domain compared to the more variable curves observed in Lunar Lander.

%% file: 6_conclu.tex
This work presents a novel approach for generating counterfactual explanations in continuous action RL. By extending the TD3 algorithm with mechanisms for constraint handling and reward shaping, our approach efficiently generates counterfactual trajectories that improve outcomes while minimizing deviations from observed actions. Experiments in the Diabetes Control and Lunar Lander domains demonstrate our approach's effectiveness, efficiency, and generalization across diverse environments. Notably, the P1 variant (without constrained states) achieves the best overall performance, while the P2 variants (with constrained states) offer flexibility for incorporating user-defined constraints and preferences.

Future directions include enhancing the interpretability of counterfactuals through real-time feedback and visualization tools, further improving user trust and understanding. Additionally, we plan to apply the approach in other high-stakes domains, such as autonomous driving, to assess its broader applicability and potential impact.